\documentclass[11pt,a4paper]{article}
\usepackage[utf8]{inputenc}

\usepackage[hyperref]{acl2018}
\usepackage{times}
\usepackage{latexsym}

\usepackage{multirow}

\usepackage{url}

\aclfinalcopy 
\def\aclpaperid{282}

\title{Embedding Learning Through Multilingual Concept Induction}

\author{Philipp Dufter\textsuperscript{1}, Mengjie Zhao\textsuperscript{2}, Martin Schmitt\textsuperscript{1}, Alexander
 Fraser\textsuperscript{1}, Hinrich Sch\"{u}tze\textsuperscript{1}\\
\textsuperscript{1} Center for Information and Language Processing (CIS) LMU Munich, Germany\\
\textsuperscript{2} École Polytechnique Fédérale de Lausanne, Switzerland\\
 {\tt \{philipp,martin,fraser\}@cis.lmu.de, mengjie.zhao@epfl.ch}}

\date{}

\def\dnrm#1{\mbox{$_{\hbox{\scriptsize #1}}$}}

\usepackage{amsfonts}
\usepackage{bm}
\usepackage{array,amsmath}
\usepackage{dsfont}
\usepackage{float}
\usepackage{graphicx}
\usepackage{algorithm}
\usepackage{algorithmicx}
\usepackage{algpseudocode}
\usepackage{pgf,tikz}
\usepackage{mathrsfs}
\usetikzlibrary{arrows}

\def\figref#1{Figure~\ref{fig:#1}}
\def\figlabel#1{\label{fig:#1}\label{p:#1}}

\def\tabref#1{Table~\ref{tab:#1}}
\def\tablabel#1{\label{tab:#1}\label{p:#1}}

\def\eqref#1{Eq.~\ref{eqn:#1}}

\newcounter{notecounter}
\newcommand{\enotesoff}{\long\gdef\enote##1##2{}}
\newcommand{\enoteson}{\long\gdef\enote##1##2{{
\stepcounter{notecounter}
{\large\bf
\hspace{1cm}\arabic{notecounter} $<<<$ ##1: ##2
$>>>$\hspace{1cm}}}}}
\enoteson
\enotesoff

\newcounter{lowerprioritynotecounter}
\newcommand{\elowerprioritynotesoff}{\long\gdef\elowerprioritynote##1##2{}}
\newcommand{\elowerprioritynoteson}{\long\gdef\elowerprioritynote##1##2{{
\stepcounter{lowerprioritynotecounter}
{\large\bf
\hspace{1cm}\arabic{lowerprioritynotecounter} $<<<$ ##1: ##2
$>>>$\hspace{1cm}}}}}
\elowerprioritynoteson
\elowerprioritynotesoff

\def\mathlinebreak{\\[0.1cm]}
\def\mathindent{\mbox{\hspace{0.5cm}}}

\begin{document}
\maketitle

\begin{abstract}
We present a new method for estimating vector space
representations of words: embedding learning by concept
induction. We
test this method on a highly parallel corpus and learn
semantic representations of words in
1259 different
languages in a single common space. An extensive
experimental evaluation on crosslingual word similarity and sentiment
analysis indicates that concept-based multilingual embedding
learning performs better than previous approaches.
\end{abstract}

\enote{as}{
	
	\begin{itemize}
		
		\item flowchart
		\item some real evaluation on a subset of languages
		\item monolingual baseline is missing
		\item typological application,
		but typoloigcail evaluation is  missing
		
	\end{itemize}
	
}


\section{Introduction}
Vector space representations of words are widely used 
because they improve performance on monolingual tasks.  This
success has generated interest in multilingual embeddings,
shared representation of words across languages
\cite{klementiev2012inducing}.  Such embeddings can be beneficial in machine
translation in sparse data settings because multilingual
embeddings provide meaning representations of source and
target in the same space. Similarly, in transfer learning,
models trained in one language on multilingual embeddings
can be deployed in other languages
\cite{zeman08crosslanguage,mcdonald11delexicalized,tsvetkov14metaphor}.
Automatically learned embeddings have the added advantage of
requiring fewer resources for training
\cite{klementiev2012inducing,hermann2014multilingual,guo16transfer}.
Thus,
massively multilingual word embeddings (i.e., covering
100s or 1000s of languages) are likely to be important  in
NLP.

The basic information many embedding learners use
is \emph{word-context information}; e.g., the embedding of a word is
optimized to predict a representation of its context. 
We instead learn embeddings from \emph{word-concept
information}. As a first approximation, a concept is a
set of semantically similar words. \figref{exclique} shows an
example concept
and also indicates one way we learn concepts:
\emph{we interpret
cliques in
the dictionary graph as concepts}. 
The nodes
of the dictionary graph
are words, its edges connect
words that are translations of each other.
A dictionary node
has the form prefix:word, e.g., ``tpi:wara'' (upper
left node in the figure).
The prefix is the ISO 639-3 code of the language; tpi is Tok Pisin.

\begin{figure}
  \centering
\includegraphics[width=0.35\textwidth]{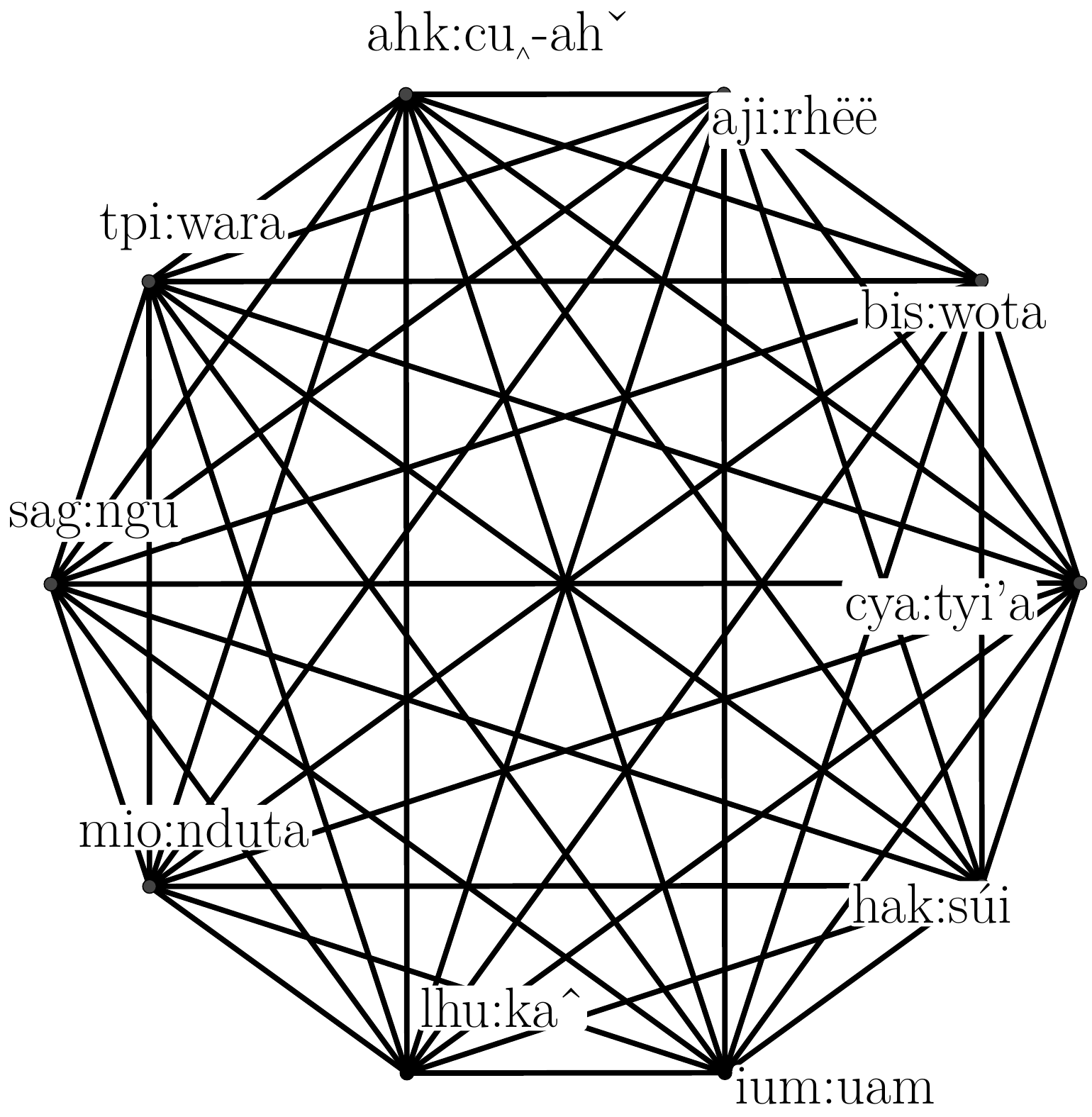}
\caption{Example of a CLIQUE concept: ``water''\figlabel{exclique}}
\end{figure}

Our method takes a parallel corpus as input and induces
a dictionary graph from the parallel corpus.  Concepts and
word-concept pairs are then induced from the
dictionary graph. Finally, embeddings are learned from
word-concept pairs.

\begin{table*}
	\scriptsize
        \begin{tabular}{p{3.3cm}@{\hspace{0.3cm}}p{6cm}@{\hspace{0.3cm}}p{5.6cm}}
		\multicolumn{1}{c}{English King James Version (KJV)} & \multicolumn{1}{c}{German Elberfelder 1905} & \multicolumn{1}{c}{Spanish Americas} \\
		\hline
		And he said , Do it the second time . And
                they did it the second time \ldots
                &Und er sprach : Füllet vier
                Eimer mit Wasser , und gießet es auf das
                Brandopfer und auf das Holz . Und er sprach
                : Tut es zum zweiten Male ! Und sie taten es
                zum zweiten Male \ldots
                &Y dijo : Llenad cuatro
                cántaros de agua y derramadla sobre el
                holocausto y sobre la leña . Después dijo :
                Hacedlo por segunda vez ; y lo hicieron por
                segunda vez \ldots
	\end{tabular}
	\caption{Instances of verse 11018034. This
          multi-sentence verse is an example of verse misalignment.}
	 \tablabel{myexverses}
\end{table*}

A key application of multilingual embeddings is transfer
learning. Transfer learning is mainly of interest if the
target is resource-poor. We therefore select as our
dataset
1664 translations in 1259 languages of the New Testament from
PBC, the Parallel Bible Corpus.
Since ``translation'' is an ambiguous word, we will from now
on refer to the 1664 translations as ``editions''.
PBC is aligned on the verse
level; most verses consist of a single sentence, but some
contain several (see \tabref{myexverses}). PBC is a good model for
resource-poverty; e.g., the training set (see below)  of KJV contains fewer than 150,000
tokens in 6458 verses.

We evaluate multilingual embeddings on two tasks, roundtrip
translation (RT) and sentiment analysis.  RT
on the word level is -- to our knowledge -- a novel
evaluation method: a query word $w$ of language $L_1$ is
translated to its closest (with respect to embedding
similarity) neighbor $v$ in $L_2$ and then backtranslated to
its closest neighbor $w'$ in $L_1$. RT is
successful if $w=w'$. There are well-known concerns about
RT when it is used in the context of
machine translation. A successful roundtrip translation does
not necessarily imply that $v$ is of high quality and it is
not possible to decide whether an error occurred in the
forward or backward translations. Despite these concerns
about RT on the sentence level, we show
that RT on the word level is a difficult task and an effective
measure of embedding quality.

\textbf{Contributions.}
(i) We introduce a new embedding
    learning method, multilingual embedding
    learning through concept induction.
(ii) We show that this new concept-based method outperforms
  previous approaches to multilingual embeddings.
  (iii) We propose both word-level and character-level
    dictionary induction methods and present evidence that
    concepts
    induced from word-level dictionaries are better for 
easily tokenizable languages and
    concepts
    induced from character-level dictionaries are better for
    difficult-to-tokenize languages.
  (iv) We evaluate our methods on a corpus of
  1664 editions in 1259 languages.
To the best of our knowledge,
this is the first detailed evaluation, involving challenging
tasks like word translation and crosslingual sentiment analysis,
that has been
    done on such a large number of languages.

\section{Methods}

\subsection{Pivot languages}

Most of our methods are based on bilingual dictionary
graphs. With 1664 editions, it is computationally expensive
to consider all editions simultaneously (more than $10^6$
dictionaries). Thus we split the set of editions in 10 pivot
and 1654 remaining editions, and do not compute
nor use dictionaries within the 1654 editions. We refer to the ten pivot editions as \emph{pivot
	languages} and give them a distinct role in concept induction.
We refer to all editions (including  pivot editions)
as \emph{target editions}. Thus, a pivot edition
has two roles:
as a pivot language and as a target edition.

We select the pivot languages based on their sparseness. Sparseness is a challenge in NLP. In the case of embeddings,
it is hard to learn a high-quality embedding for
any infrequent
word.  Many of the
world's languages (including many PBC languages) exhibit a high degree of sparseness.
But  some languages suffer
comparatively little from sparseness when simple
preprocessing
like downcasing and splitting on whitespace
is employed.

A simple measure of sparseness that affects embedding
learning is the number of types. 
Fewer types is better since their average frequency will be higher.
\tabref{pivotlangs} shows the ten languages in PBC that have
the smallest number of types in 5000 randomly selected
verses.
We randomly sample 5000 verses per edition and compare the
number of types based on this selection because
most editions do not contain a few of the selected 6458 verses.

\subsection{Character-level modeling (CHAR)}
We will see that tokenization-based models have poor
performance on a subset of the 1259 languages.  To
overcome tokenization problems, we represent a verse of
length $m$ bytes, as a sequence of $m$ $-$ $(n$ $-$ $1)$ $+$
2 overlapping byte $n$-grams. In this paper, ``$n$-gram''
always refers to ``byte $n$-gram''.  We pad the verse with
initial and final space, resulting in two additional
$n$-grams (hence ``+2''). This representation is in the
spirit of earlier byte-level processing, e.g.,
\cite{gillick2016multilingual}. There are several
motivations for this. (i) We can take advantage of
byte-level generalizations. (ii) This is robust if there is
noise in the byte encoding. (iii) Characters have different
properties in different languages and encodings, e.g.,
English UTF-8 has properties different from Chinese
UTF-8. Thus, universal language processing is easier to
design on the byte level.

We refer to this ngram representation as CHAR and
to standard tokenization as WORD.


\def\pivotsep{0.075cm}

\begin{table}
  \footnotesize
  \begin{tabular}{l@{\hspace{\pivotsep}}l@{\hspace{\pivotsep}}l@{\hspace{\pivotsep}}r@{\hspace{0.15cm}}c}
    iso&name&family; (example) region&\multicolumn{1}{c}{\rotatebox{90}{types}} & \multicolumn{1}{l}{\hspace{-0.15cm}\rotatebox{90}{tokens}} \\\hline
    lhu&Lahu&Sino-Tibetan; Thailand   &1452&268\\
ahk&Akha&Sino-Tibetan; China  &1550&315\\
hak&Hakka Chinese&Chinese; China &1596&242\\
ium&Iu Mien&Hmong-Mien; Laos   &1779&191\\
tpi&Tok Pisin&Creole; PNG&1815&177\\
mio&Pinotepa Mixtec&Oto-Manguean; Oaxaca&1828&208\\
cya&Highland Chatino &Oto-Manguean; Oaxaca&1868&231\\
bis&Bislama&Creole; Vanuatu&1872&226\\
aji&Aji\"{e}&Austronesian;
Houa\"{i}lou
&1876&194\\
sag&Sango&Creole; Central Africa&1895&192
\end{tabular}
\caption{\tablabel{pivotlangs}Our ten pivot languages, the
  languages in PBC with the lowest number of types. Tokens
  in 1000s.
Tok Pisin and 
Bislama are English-based
and Sango is a Ngbandi-based creole.
PNG = Papua New Guinea}
\end{table}

\begin{figure}[t]
	\begin{algorithm}[H]
		  \small
		\algrenewcommand\algorithmicindent{0.4cm}
		\caption{\small $\chi^2$-based dictionary induction}
		\begin{algorithmic}[1]
			\Procedure{DictionaryGraph}{$C$}
			\State $A = \text{all-edges}(C)$, $E = []$
			\For {$d \in [1,2,\ldots, d\dnrm{max}]$}
			\State $f\dnrm{max} = 2$
			\While {$f\dnrm{max} \leq |C|$}
			\State $f\dnrm{min} = \max(\min(5,f\dnrm{max}),\, \frac{1}{10}f\dnrm{max})$
			\State $(\chi^2,s,t) = \text{max-}\chi^2\text{-edge}(A,f\dnrm{min},f\dnrm{max},d)$
			\If {$\chi^2<\chi\dnrm{min}$}
			\State $f\dnrm{max} = f\dnrm{max} +1$; continue
			\EndIf
			\State $T= \mbox{extend-ngram}(A,f\dnrm{min},f\dnrm{max},d,s,t)$
			\State $\mbox{append}(E,s,T)$
			\State $\mbox{remove-edges}(A,s,T)$
			\EndWhile
			\EndFor
			\State \Return $\text{dictionary-graph}=(\text{nodes}(E), E$)
			\EndProcedure
		\end{algorithmic}
	\end{algorithm}

\caption{$\chi^2$-based dictionary induction. $C$ is a sentence-aligned corpus.
$A$ is initialized to contain all edges, i.e., the fully connected
  bipartite graph for each parallel verse.
  $E$ collects the selected dictionary edges.
 $d$ is the edge degree: in each pass
  through the loop only edges are considered whose
  participating units have a degree less than $d$.
$f\dnrm{max}$ is
  the maximum frequency during this pass.
$|C|$ is the number of sentences
  in the corpus. extend-ngram extends a target ngram to 
  left / right; e.g., if $s=$ ``jisas'' is aligned with
  ngram $t=$ ``Jesu'' in English, then ``esus'' is added
  to $T$. $t$ is always a member of $T$. remove-edges
  removes edges in $A$ between $s$ and a member of $T$.
  \figlabel{gregal}}
\end{figure}

\subsection{Dictionary induction}

\textbf{Alignment-based dictionary.}
We use fastalign \cite{dyer2013simple} to compute word alignments and use GDFA for symmetrization.
All alignment edges that
occurred at least twice
are added to the dictionary graph. Initial experiments
indicated that alignment-based dictionaries have poor quality for CHAR, probably due to the fact that overlapping
ngram representations of sentences have properties quite
different from the tokenized sentences that aligners are
optimized for. Thus we use this dictionary induction method only for WORD and developed the following alternative  for CHAR.

\textbf{Correlation-based dictionary ($\chi^2$).}
$\chi^2$ is a greedy algorithm, shown in \figref{gregal}, that
selects, in each iteration, the pair of units that has the highest
$\chi^2$ score for cooccurrence in verses. Each selected
pair is added to the dictionary and removed from the corpus.
Low-frequency units are selected first and high-frequency
units last; this prevents errors due to spurious association of
high-frequency units with low-frequency units. We perform
$d\dnrm{max}=5$ passes; in each pass, the maximum degree
of a
dictionary node is  \mbox{$1
\leq d \leq d\dnrm{max}$}. So if the node has reached degree
$d$, it is ineligible for additional edges during this
pass.
Again, this avoids errors due to spurious association
of high-frequency units
that already participate in many edges 
with low-frequency units. Recall that this method is only applied for CHAR.

\textbf{Intra-pivot dictionary.}
We assume that pivot languages are easily tokenizable. Thus we only consider alignment-based dictionaries (in total 45) within the set of pivot languages. 

\textbf{Pivot-to-target dictionary.}  We compute an
alignment-based and a $\chi^2$-based dictionary between each
pivot language and each target edition, yielding a total of
10*1664 dictionaries per dictionary type. (Note that this
implies that, for $\chi^2$,
the WORD version of the pivot
language is aligned with its CHAR version.)

\begin{figure}
	\begin{algorithm}[H]
		\small
		\algrenewcommand\algorithmicindent{0.4cm}
		\caption{\small CLIQUE concept induction}
		\begin{algorithmic}[1]
			\Procedure{Concepts}{$I\in \mathbb{R}^{n\times n}$, $\theta$, $\nu$}
			\State $G=([n],\,\{(i, j) \in [n]\times[n]\;|\; I_{ij} > \theta \})$
			\State $\text{cliques}=\text{get-cliques}(G, 3)$
			\State $G_c:=(V_c,E_c)=(\emptyset, \emptyset)$
			\For {$c_1, c_2\in \text{cliques}\times \text{cliques}$}
			\If {$|c_1 \cap c_2| \geq \nu\min\{|c_1|, |c_2|\}$}
			\State $V_c = V_c \cup \{c_1, c_2\}$, $E_c =E_c \cup \{(c_1, c_2)\}$
			\EndIf
			\EndFor
			\State $\text{metacliques}=\text{get\_cliques}(G_c, 1)$
			\State $\text{concepts} = \{\text{flatten}(c) \,|\, c \in \text{metacliques}\}$
			\State \Return concepts
			\EndProcedure
		\end{algorithmic}
	\end{algorithm}
	\vspace{-0.5cm}
	\caption{CLIQUE concept induction. $I$ is a normalized 
		adjacency matrix of a dictionary graph (i.e., relative frequency of alignment edges with respect to possible alignment edges).
          \mbox{get-cliques$(G,n)$} returns all cliques in $G$ of
          size greater or equal to $n$. \mbox{flatten$(A)$}
          flattens a set of sets.
$[n]$ denotes $\{1,2,\dots,n\}$. $\theta=0.4$, $\nu=0.6$.\figlabel{concept_ident}}
\end{figure}

\subsection{Concepts}
A concept is defined as a set of units
that has two subsets: (i) a defining set of words
from the ten pivot
languages and (ii) a set of target units (words or $n$-grams) that
are linked, via dictionary edges, to the pivot subset.
We selected the
ten ``easiest'' of the 1664 editions as
pivot languages.
Our premise is 
that semantic information is encoded
in a simply accessible form in the pivot languages and so they
should offer a good basis for learning concepts.

We induce concepts from 
the dictionary graph, a multipartite graph consisting of ten
pivot language node/word sets and
all target edition 
node/unit sets (where units are words or $n$-grams).
Edges either connect pivot nodes with other pivot nodes or pivot
nodes with target units.


\subsubsection{CLIQUE concept induction}
If concepts corresponded to each other in the overtly coding
pivot languages, if words were not ambiguous and if alignments were perfect, then
concepts would be cliques in the pivot part of the dictionary graph. These conditions are 
too strict for natural languages, so we relax them in 
our CLIQUE concept induction algorithm (\figref{concept_ident}).
The algorithm
identifies maximal multilingual cliques \mbox{(size $\geq 3$)} within the dictionary graph of
the pivot languages and then merges two cliques if they
share enough
common words. The merging lets us  identify clique-based concepts
even if, e.g., 
a dictionary edge between two words is missing. It also
accommodates the situation where more than one word of a pivot language should be
part of a concept. The merging step can also be interpreted as metaconcept induction.

Once we have identified the cliques, we project them to the
target editions: a target-unit is added to a clique if it is
connected to a proportion $\nu=0.6$ of its member words (to allow for
missing edges). This identifies around 150k clique
concepts that cover around 8k of the total vocabulary of 24k
English words
(WORD).



As an alternative to cliques,
\newcite{ammar2016massively} use
connected components (CCs). The reachability relation
(induced by CC) is the transitive closure of the edge
relation. This results in semantically unrelated words being
in the same concept for very low levels of noise. In
contrast, cliques are more ``strict'': only node subsets are
considered whose corresponding edge relation is already
transitive (or almost so for $\nu=0.6$).
Transitivity across
languages often does not hold in alignments or dictionaries;
see, e.g., 
\newcite{simard1999text}.
This
is why we only consider cliques (which reflect already
existent transitivity) rather than CCs, which impose
transitivity where it does not hold naturally.

\elowerprioritynote{pd}{perhaps reuse "A
	problem with CCs is
	that one typically  gets one very large CC
	spanning almost the full vocabulary) and only a few of moderate size. The number of CCs was too low for training embeddings."}

\begin{figure}
	\centering
	\scriptsize
\begin{tabular}{p{0.5\textwidth}}
$N(t)=$\{\texttt{bis:Jorim, ium:yo-lim, sag:Yorim, tpi:Jorim}\}\\\\
$t$$\in$$T$$=$\{\texttt{ac0:Yorim,atg0:iJorimu,bav0:Jorim,bom0:Yorim, dik0:Jorim, dtp0:Yorim, duo0:Jorim, eng1:Jorim, engb:Jorim, fij2:Lorima, fij3:Jorima, gor0:Yorim, hvn0:Yorim, ibo0:Jorim, iri0:Jorri, kmr0:Yorîm, ksd0:Iorim, kwd0:Jorim, lia0:Yorimi, loz0:Jorimi, mbd0:Hurim, mfh0:Yorim, min0:Yorim, mrw0:Yorim,mse0:Jorimma,naq0:Jorimmi, smo1:Iorimo, srn1:Yorim, tsn2:Jorime, yor2:Jórímù}\}
\end{tabular}
\caption{Target neighborhood concept example: $N(t) \cup T$. $N(t)$ is the target neighborhood for each of the target
	words in $T$.
	\figlabel{concex}}
\end{figure}

\subsubsection{$N(t)$ (target neighborhood) concept induction}
Let $N(t)$ be the neighborhood of target node
$t$ in the multipartite dictionary graph, i.e., the set of
pivot words that are linked to $t$. We refer to $N(t)$ as \emph{target neighborhood}.
\figref{concex} shows an example of such a target neighborhood,
the set $N(t)$ consisting of four words.\footnote{We use numbers and lowercase letters
		at the fourth position of the prefix
		to
		distinguish different editions in the same language, e.g.,
		``0'', ``3'' and ``e'' in
		``ace0'',
		``fij3'', ``enge''.}
A \emph{target neighborhood concept} consists of a set $T$ of pivot
words and all target words $t$ for which $T=N(t)$ holds.

\textbf{Motivation.} Suppose
$N(t) = N(u)$ for target nodes $t$ and $u$ from two different
languages and $|N(t)|$ covers several pivot languages, e.g., 
$|N(t)|=
|N(u)|=4$ as in the figure. 
Again, if units closely corresponded to concepts,
if there were no ambiguity, if the dictionary were perfect,
then we could safely conclude that the meanings of $t$ and
$u$ are  similar; if the meanings of $t$ and $u$ were
unrelated,
it is unlikely that they
would be aligned to the exact same words
in four different languages. In reality, there is no exact
meaning-form correspondence, there is ambiguity and the
dictionary is not perfect. Still, we will see below that
defining concepts as target neighborhoods works well.

\subsubsection{Filtering target neighborhood concepts}
In contrast to CLIQUE, we do not put any
constraint on the pivot-to-pivot connections within target
neighborhoods; e.g., in \figref{concex}, we do not require
that ``bis:Jorim'' and
``sag:Yorim''  are connected by an edge.
We evaluate three
postfiltering steps of target neighborhoods to increase their
quality: 
restricting target
neighborhoods to those that are cliques
in \textbf{$N(t)$-CLIQUE};
to those that are
connected components
in
\textbf{$N(t)$-CC}; and to those 
of
size two that are valid edges in
the  dictionary
in \textbf{$N(t)$-EDGE}.
For \textbf{$N(t)$-EDGE},
we found that
taking all edges performs well, so we also consider edges
that are proper subsets of target neighborhoods.

\subsection{Embedding learning}
We adopt the framework of
embedding learning algorithms that define contexts and then
sample pairs of an input word (more generally, an input unit) and a context word
(more generally, a context unit) from each context.
The only difference is that our contexts are concepts.
For simplicity, we use word2vec \cite{mikolov2013efficient}
as the implementation of
this model.\footnote{We use {\scriptsize \url{code.google.com/archive/p/word2vec}}}

\subsection{Baselines}
Baselines
for \textbf{multilingual embedding learning.} One baseline is inspired by \cite{vulic2015bilingual}. We consider
words of one aligned verse in the pivot languages and one target language as a bag of words (BOW) and consider this bag as
a context.\footnote{The actual implementation slightly differs to avoid very long lines. It does only consider two pivot languages at a time, but writes each verse multiple times.}

\newcite{levy2017strong} show that sentence ID features
(interpretable as an abstract representation of the word's
context) are effective. We use a corpus with lines
consisting of pairs of an identifier of a verse and a unit 
extracted from that verse as input to word2vec and call this baseline S-ID.

\newcite{lardilleux2009sampling} 
propose a simple and efficient
baseline:
\textbf{sample-based concept induction}. Words that strictly occur in the same verses are assigned to the same concept. To increase coverage, they propose to sample many different subcorpora.\footnote{We use this implementation:  {\scriptsize \url{anymalign.limsi.fr}}} We induce concepts using this method and project them analogous to CLIQUE. We call this baseline SAMPLE.

One novel contribution of this paper is \textbf{roundtrip evaluation}
of embeddings.  We learn embeddings based on a 
dictionary. The question arises: are the embeddings simply
reproducing the information already in the dictionary or are
they improving the performance of roundtrip search?

As a baseline, we perform
RTSIMPLE, a simple dictionary-based
roundtrip translation method. Retrieve the pivot word $p$ in
pivot language $L_p$ (i.e., $p \in L_p$) that is closest to the query
$q \in L_q$. Retrieve the target unit $t \in L_t$
that is closest to $p$. Retrieve the 
pivot word $p' \in
L_p$ that is closest to $t$.
Retrieve the unit $q' \in L_q$
that is closest to $p'$. If $q=q'$, this is an exact hit. We
run this experiment for all pivot and target languages. 

Note that roundtrip evaluation tests the capability of a
system to go from any language to any other language. In an
embedding space, this requires two hops. In a highly
multilingual dataset of
$n$ languages in which not all $O(n^2)$ bilingual
dictionaries exist, this requires four hops.

\section{Experiments and results}
\subsection{Data}
We use PBC
\cite{mayer2014creating}. The version we pulled on
2017-12-11 contains 1664 Bible editions  in
1259 languages (based on ISO 639-3 codes) after we discarded
editions that have low coverage of the New Testament.
We use 7958 verses that have good coverage in these 1664 editions.
The data is verse
aligned; a verse of the New Testament can consist of
multiple sentences.
We randomly split verses
6458/1500
into train/test.

\subsection{Evaluation}
For \textbf{sentiment analysis}, we represent a verse as the
IDF-weighted sum of its embeddings. Sentiment classifiers
(linear SVMs) are trained on the training set
of the
World English Bible edition
for the
two decision problems positive vs.\ non-positive and
negative vs.\ non-negative. We
create a silver standard by labeling verses in English
editions with the NLTK
\cite{bird2009natural}  sentiment classifier.

A positive vs.\ negative classification is not reasonable
for the New Testament because a large number of verses is
mixed, e.g.,
``Now is come salvation \ldots\ 
the power of his Christ: for the accuser \ldots\ cast down, which accused them before our
God \ldots''
Note that this verse also cannot be said to be neutral.  Splitting
the sentiment analysis into two subtasks (``contains
positive sentiment: yes/no'' and ``contains negative
sentiment: yes/no'') is
an effective solution for this paper.

The two trained models are then applied to the test set of
all 1664 editions. All embeddings in this paper are learned on the training set
only. So no test information was used for learning the embeddings.

\textbf{Roundtrip translation.}
There are no gold standards for
the genre of our corpus (the New Testament); for only a few
languages out-of-domain gold standards are available.
Roundtrip evaluation is an evaluation method for multilingual
embeddings that can be applied if no resources are available
for a language. Loosely speaking, for a query $q$ in a query language $L_q$ (in our case English) and a target
language $L_t$, roundtrip translation finds the unit $w_t$ in $L_t$ that is
closest to $q$ and then the English unit $w_e$ that is
closest to $w_t$. If the semantics of $q$ and $w_e$ are
identical (resp.\ are unrelated), this is deemed evidence for
(resp.\ counter-evidence against)
the quality of the
embeddings.
We work on the level of Bible edition, i.e., two editions in
the same language are considered different ``languages''.

For a query $q$, we denote the set of its $k_I$ nearest
neighbors in the target edition $e$ by $I_e(q)=\{u_1, u_2,
\dots, u_{k_I}\}$. For each intermediate entry we then
consider its $k_T$ nearest neighbors in English. Overall we
get a set $T_{e}(q)$ with $k_I k_T$ predictions for each
intermediate Bible edition $e$. See \figref{exrttspa} for an
example.

We evaluate the predictions $T_e(q)$ using two sets $G_s(q)$
(strict) and $G_r(q)$ (relaxed) of ground-truth semantic
equivalences in English.
Precision for a
query $q$ is defined as\mathlinebreak
\mathindent $p_i(q) := 1/|E|\sum_{e \in E} \min\{1, |T_e(q) \cap G_i(q)|\}$\mathlinebreak
where $E$ is the set of all Bible editions and 
$i\in\{s,r\}$. We report the mean and median across a set of $70$ queries selected from
\citet{swadesh1971south}'s  list of 100 universal linguistic
concepts.  

We create $G_s$ and $G_r$ as follows.
For WORD, we  define $G_s(q)=\{q\}$ and $G_r(q) = L(q)$ where
$L(q)$ is the set of words with the same lemma and POS as $q$.  For
CHAR, we need to find ngrams that correspond uniquely to the
query $q$. Given a candidate ngram $g$ we consider
$c_{qg}:=1/c(g)\sum_{q' \in L(q), \mbox{substring}(g,q')}c(q')$
where $c(x)$ is the count of  character sequence $x$
across all editions in the query language. We
add $g$ to $G_i(q)$ if  $c_{qg}> \sigma_i$
where $\sigma_s =.75$ and $\sigma_r = .5$. We only consider queries where $G_s(q)$ is non-empty.

We vary the evaluation parameters $(i, k_I, k_T)$ as follows: 
``S1'' represents $(s, 1,1)$,
``S4'' $(s, 2,2)$,
``S16'' $(s, 2,8)$,
and
``R1''
$(r, 1,1)$.

\def\sepmujer{0.08cm}

\begin{figure}
	\centering
	\scriptsize
	\begin{tabular}{c@{\hspace{\sepmujer}}c@{\hspace{\sepmujer}}c@{\hspace{\sepmujer}}c@{\hspace{\sepmujer}}p{4.5cm}}
		 & & inter- & & \\
		query & & mediate & & predictions\\
		\hline
        \texttt{woman} & $\Rightarrow$ & \texttt{mujer} & $\Rightarrow$ & \texttt{wife} \texttt{woman} \texttt{women} \texttt{widows} \texttt{daughters} \texttt{daughter} \texttt{marry} \texttt{married} \\
        & $\Rightarrow$ & \texttt{esposa} & $\Rightarrow$ & \texttt{marry} \texttt{wife} \texttt{woman} \texttt{married} \texttt{marriage} \texttt{virgin} \texttt{daughters} \texttt{bridegroom} \\
 	\end{tabular}
 	\caption{Roundtrip translation example for KJV
 		and Americas Bible (Spanish). In this example $\min\{1, |T_e(q) \cap G_i(q)|\}$ equals $0$ for S1 and R1, and $1$ for S4 and S16.\figlabel{exrttspa}}
\end{figure}

\begin{figure}
	\scriptsize
	\begin{tabular}{p{7cm}}
\texttt{connu(3), connais(3), connaissent(3), savez(2), sachant(2), sait(2), sachiez(2), savoir, s\c{c}ai, ignorez, connaissiez, sache connaissez, connaissais, savent, savaient, connoissez, connue, reconnaîtrez, sais, connaissant, savons, connaissait, savait}
	\end{tabular}
\caption{Intermediates  aggregated over  17
	French editions. $q$=``know'',
	$N(t)$ embeddings, S16.
	Intermediates
	are correct with two possible exceptions:
	``ignorez'' `you do not know'; ``reconna\^itrez'' `you recognize'.
	\figlabel{exrtt}}
\end{figure}


\def\bigsep{0.1cm}

\begin{table*}
	\footnotesize
	\begin{tabular}{lr||r@{\hspace{\bigsep}}r@{\hspace{\bigsep}}r@{\hspace{\bigsep}}r@{\hspace{\bigsep}}r@{\hspace{\bigsep}}r@{\hspace{\bigsep}}r@{\hspace{\bigsep}}r@{\hspace{\bigsep}}r@{\hspace{\bigsep}}|r@{\hspace{\bigsep}}r@{\hspace{\bigsep}}r@{\hspace{\bigsep}}r@{\hspace{\bigsep}}r@{\hspace{\bigsep}}r@{\hspace{\bigsep}}r@{\hspace{\bigsep}}r@{\hspace{\bigsep}}r||r@{\hspace{\bigsep}}r|r@{\hspace{\bigsep}}r}
		&  & \multicolumn{17}{c}{roundtrip
            translation}&\multicolumn{5}{r}{sentiment analysis}\\
		&    & \multicolumn{9}{c}{WORD} & \multicolumn{9}{c}{CHAR}
		& \multicolumn{2}{c}{WORD} & \multicolumn{2}{c}{CHAR}\\\hline
		&& \multicolumn{2}{c}{S1} & \multicolumn{2}{c}{R1} & \multicolumn{2}{c}{S4} & \multicolumn{2}{c}{S16}& 
		& \multicolumn{2}{c}{S1} & \multicolumn{2}{c}{R1} & \multicolumn{2}{c}{S4} & \multicolumn{2}{c}{S16}& \\
		&    & $\mu$ &  Md &$\mu$ & Md &$\mu$ & Md &$\mu$ & Md &N
		& $\mu$ & Md &$\mu$ & Md &$\mu$ & Md &$\mu$ & Md &N&\rotatebox{01}{pos} & \rotatebox{01}{neg} &\rotatebox{01}{pos} & \rotatebox{01}{neg} \\\hline\hline
1&RTSIMPLE&33&24&37&36&&&&&67&24&13&32&21&&&&&70  &&&&\\
2&BOW&7&5&8&7&13&12&26&28&69&3&2&3&2&5&4&10&11&70 &33&81&13&83\\
3&S-ID&46&46&52&55&63&76&79&91&65&9&5&9&5&14&9&25&22&70 &79&88&65&86\\
4&SAMPLE&33&23&43&42&54&59&82&96&65&53&\textbf{59}&59&\textbf{72}&67&85&79&99&58 &82&89&77&89\\
5&CLIQUE&43&36&59&63&67&77&93&99&69&42&46&48&55&60&76&73&98&53 &84&89&69&88\\
6&N(t)&\textbf{54}&\textbf{59}&\textbf{61}&69&\textbf{80}&\textbf{87}&\textbf{94}&\textbf{100}&69&50&53&54&59&73&82&90&99&66 &82&89&\textbf{87}&\textbf{90}\\
7&N(t)-CC&52&56&59&66&77&86&93&99&69&40&45&42&48&58&69&75&95&57& 80&88&68&86\\
8&N(t)-CLIQUE&11&0&11&0&16&0&22&0&18&39&45&41&47&58&74&76&94&56 &22&84&61&84\\
9&N(t)-EDGE&35&30&43&36&56&55&87&94&69&39&29&49&52&64&78&88&\textbf{100}&63 &84&\textbf{90}&84&89
	\end{tabular}
	
	\caption{Roundtrip translation (mean/median accuracy) and sentiment
		analysis ($F_1$) results for word-based (WORD) and character-based (CHAR)
		multilingual embeddings. $N$ (coverage): \#
		queries  contained in the embedding space.
		The best result \emph{across WORD and CHAR} is set in
		bold.}
	\tablabel{bigtable}
\end{table*}

\subsection{Corpus generation and hyperparameters}
 We
train with the 
skipgram model and set vector dimensionality to 200;
word2vec default parameters are used otherwise.
Each concept -- the union of a set of pivot words and a set
of target units linked to the pivot words -- is written out
as a line or (if the set is large) as a sequence of shorter
lines.  Training corpus size is approximately 50 GB for all
experiments.  We write several copies of each line (shuffling randomly to ensure lines are different)
where the multiplication factor is chosen to
result in an overall corpus size of approximately 50 GB.

There are two exceptions. For BOW, we did not find a good way
of reducing the corpus size, so this corpus is 10 times
larger than the others. For S-ID, we use
\newcite{levy2017strong}'s hyperparameters; in particular, we trained for 100
iterations and we wrote each verse-unit pair to the corpus
only once, resulting in a corpus of about 4 GB.

We
set the $n$ parameter of $n$-grams to
$n=4$ for
Bible editions with $\rho<2$, $n=8$ for
Bible editions with $2 \leq \rho<3$ and $n=12$ for
Bible editions
with $\rho \geq 3$ where $\rho$ is the ratio between size in
bytes of the edition and median size of the 1664 editions.
In $\chi^2$ dictionary induction, we set $\chi\dnrm{min}=100$. In the concept induction algorithm we set $\theta=0.4$ and $\nu=0.6$.
Except for SAMPLE and CLIQUE,
we filter out hapax legomena.

\subsection{Results}
\tabref{bigtable} presents evaluation results for roundtrip
translation and sentiment analysis.

\textbf{Validity of roundtrip (RT) evaluation results.} 
RTSIMPLE (line 1) is not competitive; e.g.,
its accuracy is lower by almost half compared to
$N(t)$.  We also see that RT is an excellent differentiator
of poor multilingual embeddings (e.g., BOW)
vs.\ higher-quality ones like S-ID and $N(t)$.
This indicates that RT translation can serve as an effective  evaluation
measure.

The
\textbf{concept-based multilingual embedding learning}
algorithms CLIQUE  and $N(t)$ (lines 5-6)
consistently (except S1 WORD) outperform 
BOW and S-ID (lines 2-3) that are not based
on concepts. BOW performs
poorly in our low-resource setting; this is not surprising
since BOW methods rely on large datasets and are therefore
expected to fail in the face of severe sparseness. S-ID
performs reasonably well for WORD, but even in that case it
is outperformed by $N(t)$, in some cases by a large margin,
e.g., $\mu$ of 63 for S-ID vs.\ 80 for $N(t)$ for S4.
For CHAR, S-ID results are
poor. On sentiment classification, $N(t)$ also consistently
outperforms S-ID.

While S-ID provides a clearer signal to the embedding
learner than BOW, it is still relatively crude to represent
a word as -- essentially -- its binary vector of verse
occurrence. Concept-based methods perform better because
they can exploit the more
informative  dictionary graph.

\textbf{Comparison of graph-theoretic definitions of
  concepts: $N(t)$-CLIQUE, $N(t)$-CC.}
$N(t)$ (line 6) has the most consistent good performance across tasks
and evaluation measures. Requiring target neighborhoods to be connected components
 (line 7) performs similar but does not yield any improvements.
$N(t)$-CLIQUE (line 8) does not work at all. 
The number of
target neighborhoods which are quasi-cliques is too small, resulting in 
a low number of concepts and thus a poor coverage ($N=18$). 
$N(t)$-CLIQUE
results are highly increased for CHAR, but still poorer by a
large margin than the best methods. We can interpret this result
as an instance of a precision-recall tradeoff: presumably
the quality of the concepts found by
$N(t)$-CLIQUE is better (higher precision), but there are
too few of them (low recall) to get good evaluation numbers.

\textbf{Comparison of graph-theoretic definitions of
  concepts: CLIQUE.} CLIQUE has strong performance for a
subset of measures, e.g., ranks consistently second for RT (except S1 WORD) and sentiment analysis in WORD. 
Although CLIQUE is
perhaps the most intuitive way of inducing a concept from a
dictionary graph, it may suffer in relatively
high-noise settings like ours.

\textbf{Comparison of graph-theoretic definitions of
  concepts: $N(t)$ vs.\ $N(t)$-EDGE.} 
Recall that $N(t)$-EDGE postfilters target neighborhoods by
only considering pairs of pivot words 
  that are linked by a
dictionary edge. This ``quality'' filter does seem to work
in some cases, e.g., best performance
S16 Md for CHAR. But results for WORD are much poorer.

\textbf{SAMPLE} performs best for CHAR: best results in
five out of eight cases. However, its coverage is
low: $N=58$. This is also the reason that it does not
perform well on sentiment analysis for CHAR ($F_1=77$ for
pos).


\textbf{Target neighborhoods $N(t)$.} The overall best
method is   $N(t)$. It is the best method more often than
any other method and in the other cases, it ranks second. This result
suggests that the assumption that two target units
are semantically similar if they
have
dictionary edges with exactly the same set of pivot words
is a reasonable approximation of reality. Postfiltering by
putting constraints on eligible sets of pivot words (i.e.,
the pivot words themselves must have a certain dictionary
link structure) does not consistently improve upon target neighborhoods.

\begin{figure}
	\centering
	\includegraphics[width=0.39\textwidth]{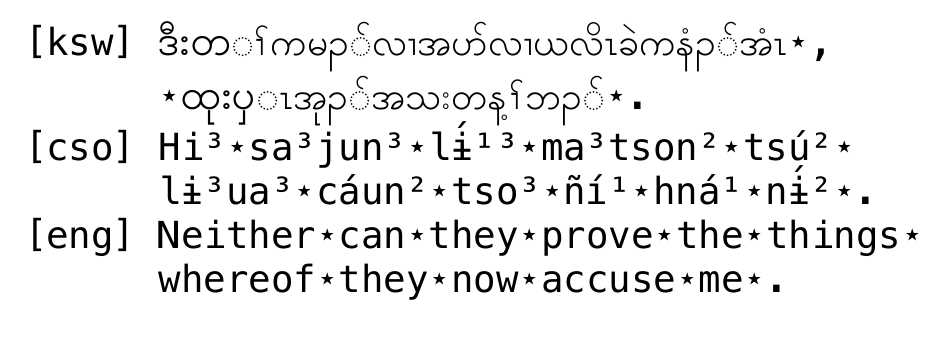}
	\vspace{-0.0cm}
	\caption{Verse 44024013. ``*'' = tokenization boundary.  S'gaw Karen (ksw) is difficult to
          tokenize and CHAR $>$ WORD for $N(t)$.
Chinanteco de
          Sochiapan (cso) has few types, similar to a pivot
          language, and CHAR $<$ WORD for $N(t)$.
                    \figlabel{charvsword}}
\end{figure}

\def\deltasep{0.15cm}

\begin{table}
	\small
	\centering
	\begin{tabular}{l@{\hspace{\deltasep}}r|l@{\hspace{\deltasep}}r|l@{\hspace{\deltasep}}r|l@{\hspace{\deltasep}}r}
		\multicolumn{2}{c}{N(t)}&\multicolumn{2}{c}{S-ID}&\multicolumn{2}{c}{SAMPLE}&\multicolumn{2}{c}{CLIQUE}\\
		\multicolumn{2}{c}{[\textsc{char}]}&\multicolumn{2}{c}{[\textsc{word}]}&\multicolumn{2}{c}{[\textsc{word}]}&\multicolumn{2}{c}{[\textsc{word}]}\\

			iso&$\Delta$&iso&$\Delta$&iso&$\Delta$&iso&$\Delta$\\
\hline
arb1 & 54 & pua0 & 61 & jpn1 & 42 & mya2 & 38 \\
arz0 & 53 & sun2 & 54 & khm2 & 40 & jpn1 & 36 \\
cop3 & 49 & jpn1 & 53 & cap2 & 40 & khm3 & 34 \\
srp0 & 44 & khm3 & 53 & khm3 & 40 & bsn0 & 28 \\
cop2 & 44 & khm2 & 50 & mya2 & 39 & khm2 & 27 \\
\dots & \dots & \dots & \dots & \dots& \dots& \dots& \dots\\
pis0 & -23 & vie7 & -24 & eng8 & -7 & haw0 & -22 \\
pcm0 & -23 & kri0 & -25 & enm1 & -9 & eng4 & -23 \\
ksw0 & -24 & tdt0 & -27 & lzh2 & -9 & enm2 & -26 \\
lzh2 & -41 & eng2 & -27 & eng4 & -12 & enm1 & -26 \\
lzh1 & -51 & vie6 & -29 & lzh1 & -13 & engj & -28 
	\end{tabular}
\vspace*{0.2 cm}
\newline
	\caption{Comparison of $N(t)$[\textsc{word}] with four
          other methods. Difference in mean performance
          (across queries) in R1 per edition. Positive number means better performance of $N(t)$[\textsc{word}]. \tablabel{deltamu}}
\end{table}

\textbf{WORD vs.\ CHAR.}
For roundtrip, WORD is a better representation than
CHAR if we just count the bold winners: seven (WORD)
vs.\ three (CHAR), with two ties. For sentiment, the more difficult task is
pos and for this task, CHAR is better by 3 points than
WORD ($F_1=87$, line 6, vs.\  $F_1=84$, lines
9/5).
However, \tabref{deltamu} shows that 
CHAR $<$ WORD
for one subset of editions
(exemplified by cso in \figref{charvsword})
and CHAR $>$ WORD
for a different subset (exemplified
by ksw).
So there are big differences between CHAR and   WORD in both
directions, depending on the language. For some languages,
WORD performs a lot better, for others, CHAR performs a lot better.

We designed RT evaluation as a word-based evaluation
that disfavors CHAR in some cases. The fourgram
``ady@'' in the World English Bible occurs in
``already'' (32 times), ``ready'' (31 times) and
``lady'' (9 times). Our RT evaluation thus disqualifies
``ady@'' as a strict match for ``ready''. But all 17 \emph{aligned}
occurrences of ``ady@'' are part of ``ready'' -- all others
were not aligned. So in the $\chi^2$-alignment interpretation,
$P(\mbox{ready}|\mbox{ady@})=1.0$. In contrast to RT, we
only used aligned ngrams in the sentiment evaluation. This
discrepancy may explain why the best method for sentiment is
a CHAR method whereas the best method for RT is a WORD method.

\textbf{First NLP task evaluation on more than 1000 languages.}
\tabref{bigtable} presents results
for   1664 editions in 1259 languages.
To the best of our knowledge,
this is the first detailed evaluation, involving two challenging
NLP tasks,
that has been
    done on such a large number of languages.
For several methods, the results are above baseline for all
1664 editions; e.g., S1 measures
 are above 20\% for all 1664
 editions for $N(t)$ on CHAR.

\elowerprioritynote{AF}{

	BTW, a new thought: I'd definitely also discuss synonomy
	in your cliques
	discussion. My thinking is that if you use something
	simpler than cliques,
	you can't capture perfect synonomy of two words in the
	same language, which
	is problematic.
	
}

\section{Related Work}



Following
\newcite{upadhyay2016cross},
we group
\textbf{multilingual embedding}  
methods  into
classes A, B, C, D. 

Group A trains monolingual embedding
spaces and subsequently uses a transformation
to create a unified space. \newcite{mikolov2013exploiting}
find the transformation by minimizing the Euclidean distance
between word pairs. Similarly, \newcite{zou2013bilingual},
\newcite{xiao2014distributed} and
\newcite{faruqui2014improving} 
use different data sources for identifying word pairs and
creating the transformation  (e.g., by CCA).
\newcite{duong2017multilingual} is also similar.
These approaches need large datasets to obtain high quality
monolingual embedding spaces and are thus 
inappropriate for a low-resource setting of 150,000 tokens per language.


Group B starts from the premise that representation of
aligned sentences should be similar.  Neural network
approaches include \cite{hermann2013multilingual} (BiCVM)
and \cite{ap2014autoencoder} (autoencoders). Again, we have
not enough data for training neural networks of this
size. \newcite{sogaard2015inverted} learn an interlingual
space by using Wikipedia articles as concepts and applying
inverted indexing. \newcite{levy2017strong} show that what
we call
S-ID  is a strongly performing 
embedding learning method. We use S-ID as a baseline.

Group C combines mono- and multilingual information in
the embedding learning objective.
\newcite{klementiev2012inducing} add a
word-alignment based term in the objective.
\newcite{luong2015bilingual} extend
\newcite{mikolov2013efficient}'s skipgram model to a
bilingual model. \newcite{gouws2015bilbowa} introduce a
crosslingual term in the objective, which does not rely on
any word-pair or alignment information.
For $n$ editions, including
 $O(n^2)$ bilingual terms in the objective function does not scale.

Group D creates pseudocorpora by merging data from
multiple languages into a single corpus.
One such method, due to \newcite{vulic2015bilingual},
is our baseline BOW.

\newcite{ostling2014bayesian} generates 
\textbf{multilingual concepts}
using a Chinese Restaurant process, a computationally
expensive method.
\newcite{wang2016novel} base their concepts on
cliques. We extend their notion of clique  from the
bilingual to the multilingual case.
\newcite{ammar2016massively} use connected components.
Our baseline SAMPLE,
based on 
\cite{lardilleux2007contribution,lardilleux2009sampling},
samples aligned sentences from a  multilingual corpus and
extracts perfect alignments.

\newcite{malaviya17typology}, \newcite{asgari17past},
\newcite{ostling2017continuous} and
\newcite{tiedemann2018emerging}
perform \textbf{evaluation} on the language
level (e.g., typology prediction) for
1000+ languages or perform experiments on 1000+ languages
without evaluating each language.
We present the first work that evaluates on 1000+ languages
on the sentence level
on a difficult task.

\newcite{somers2005round} criticizes RT evaluation
on the sentence level; but see
\newcite{aiken10roundtrip}.
We demonstrated that when used on the word/unit level, it distinguishes
weak from strong embeddings and correlates well with an
independent sentiment evaluation.

Any alignment algorithm can be used for \textbf{dictionary
  induction}. We only used a member of the IBM class of
models \cite{dyer2013simple}, but presumably we could
improve results by using either higher performing albeit
slower aligners or non-IBM aligners (e.g.,
\cite{och03alignment,tiedemann03clues,melamed97wordtoword}). Other
alignment algorithms include  2D linking
\cite{kobdani09word}, sampling based methods 
(e.g., \newcite{vulic2012sub}) and EFMARAL 
\cite{ostling2016efmaral}. EFMARAL is especially
intriguing as it is based on IBM1 and
\newcite{agic2016multilingual} find IBM2-based models to
favor closely related languages more than models based on
IBM1.  However, the challenge is that we need to compute
tens of thousands of alignments, so speed is of the
essence. We ran character-based and word-based induction
separately; combining them is promising future research;
cf.\ \cite{heyman2017bilingual}.

There is much work on embedding learning that does not require
\textbf{parallel corpora}, e.g.,
\cite{vulic2012detecting,ammar2016massively}. This work is
more generally applicable, but a parallel corpus provides a
clearer signal and is more promising (if available) for
low-resource research.

\section{Summary}
We presented a new method for estimating vector space
representations of words: embedding learning by concept
induction. We
tested this method on a highly parallel corpus and learned
semantic representations of words in
1259 different
languages in a single common space. Our extensive
experimental evaluation on crosslingual word similarity and sentiment
analysis indicates that concept-based multilingual embedding
learning performs better than previous approaches.

The embedding spaces of the 1259
languages (SAMPLE, CLIQUE and $N(t)$) are available: \\ {\footnotesize \url{http://cistern.cis.lmu.de/comult/}}.

We gratefully
  \textbf{acknowledge}
funding from
the European Research Council
(grants  740516 \& 640550) and through a
Zentrum Digitalisierung.Bayern 
fellowship awarded to
the first author. We are indebted to Michael Cysouw for
making PBC available to us.

\bibliography{acl2018}

\begin{thebibliography}{46}
\expandafter\ifx\csname natexlab\endcsname\relax\def\natexlab#1{#1}\fi

\bibitem[{Agi{\'c} et~al.(2016)Agi{\'c}, Johannsen, Plank, Mart{\'\i}nez,
  Schluter, and S{\o}gaard}]{agic2016multilingual}
{\v{Z}}eljko Agi{\'c}, Anders Johannsen, Barbara Plank, H{\'e}ctor~Alonso
  Mart{\'\i}nez, Natalie Schluter, and Anders S{\o}gaard. 2016.
\newblock Multilingual projection for parsing truly low-resource languages.
\newblock \emph{Transactions of the Association for Computational Linguistics},
  4.

\bibitem[{Aiken and Park(2010)}]{aiken10roundtrip}
Milam Aiken and Mina Park. 2010.
\newblock The efficacy of round-trip translation for {MT} evaluation.
\newblock \emph{Translation Journal}, 14(1).

\bibitem[{Ammar et~al.(2016)Ammar, Mulcaire, Tsvetkov, Lample, Dyer, and
  Smith}]{ammar2016massively}
Waleed Ammar, George Mulcaire, Yulia Tsvetkov, Guillaume Lample, Chris Dyer,
  and Noah~A Smith. 2016.
\newblock Massively multilingual word embeddings.
\newblock \emph{arXiv preprint arXiv:1602.01925}.

\bibitem[{Asgari and Sch{\"{u}}tze(2017)}]{asgari17past}
Ehsaneddin Asgari and Hinrich Sch{\"{u}}tze. 2017.
\newblock Past, present, future: {A} computational investigation of the
  typology of tense in 1000 languages.
\newblock In \emph{Proceedings of the 2017 Conference on Empirical Methods in
  Natural Language Processing}.

\bibitem[{Bird et~al.(2009)Bird, Klein, and Loper}]{bird2009natural}
Steven Bird, Ewan Klein, and Edward Loper. 2009.
\newblock \emph{Natural language processing with {P}ython: {A}nalyzing text
  with the natural language toolkit}.
\newblock O'Reilly Media.

\bibitem[{Duong et~al.(2017)Duong, Kanayama, Ma, Bird, and
  Cohn}]{duong2017multilingual}
Long Duong, Hiroshi Kanayama, Tengfei Ma, Steven Bird, and Trevor Cohn. 2017.
\newblock Multilingual training of crosslingual word embeddings.
\newblock In \emph{Proceedings of the 15th Conference of the European Chapter
  of the Association for Computational Linguistics}.

\bibitem[{Dyer et~al.(2013)Dyer, Chahuneau, and Smith}]{dyer2013simple}
Chris Dyer, Victor Chahuneau, and Noah~A Smith. 2013.
\newblock A simple, fast, and effective reparameterization of ibm model 2.
\newblock In \emph{Proceedings of the 2013 Conference of the North American
  Chapter of the Association for Computational Linguistics: Human Language
  Technologies}.

\bibitem[{Faruqui and Dyer(2014)}]{faruqui2014improving}
Manaal Faruqui and Chris Dyer. 2014.
\newblock Improving vector space word representations using multilingual
  correlation.
\newblock In \emph{Proceedings of the 14th Conference of the European Chapter
  of the Association for Computational Linguistics}.

\bibitem[{Gillick et~al.(2016)Gillick, Brunk, Vinyals, and
  Subramanya}]{gillick2016multilingual}
Dan Gillick, Cliff Brunk, Oriol Vinyals, and Amarnag Subramanya. 2016.
\newblock Multilingual language processing from bytes.
\newblock In \emph{Proceedings of the 2016 Conference of the North American
  Chapter of the Association for Computational Linguistics: Human Language
  Technologies}.

\bibitem[{Gouws et~al.(2015)Gouws, Bengio, and Corrado}]{gouws2015bilbowa}
Stephan Gouws, Yoshua Bengio, and Greg Corrado. 2015.
\newblock Bilbowa: fast bilingual distributed representations without word
  alignments.
\newblock In \emph{Proceedings of the 32nd International Conference on
  International Conference on Machine Learning}.

\bibitem[{Guo et~al.(2016)Guo, Che, Yarowsky, Wang, and Liu}]{guo16transfer}
Jiang Guo, Wanxiang Che, David Yarowsky, Haifeng Wang, and Ting Liu. 2016.
\newblock A representation learning framework for multi-source transfer
  parsing.
\newblock In \emph{Proceedings of the 30th AAAI Conference on Artificial
  Intelligence}.

\bibitem[{Hermann and Blunsom(2014{\natexlab{a}})}]{hermann2013multilingual}
Karl~Moritz Hermann and Phil Blunsom. 2014{\natexlab{a}}.
\newblock Multilingual distributed representations without word alignment.
\newblock In \emph{Proceedings of the 2014 International Conference on Learning
  Representations}.

\bibitem[{Hermann and Blunsom(2014{\natexlab{b}})}]{hermann2014multilingual}
Karl~Moritz Hermann and Phil Blunsom. 2014{\natexlab{b}}.
\newblock Multilingual models for compositional distributed semantics.
\newblock In \emph{Proceedings of the 52nd Annual Meeting of the Association
  for Computational Linguistics}.

\bibitem[{Heyman et~al.(2017)Heyman, Vuli{\'c}, and
  Moens}]{heyman2017bilingual}
Geert Heyman, Ivan Vuli{\'c}, and Marie-Francine Moens. 2017.
\newblock Bilingual lexicon induction by learning to combine word-level and
  character-level representations.
\newblock In \emph{Proceedings of the 15th Conference of the European Chapter
  of the Association for Computational Linguistics}.

\bibitem[{Klementiev et~al.(2012)Klementiev, Titov, and
  Bhattarai}]{klementiev2012inducing}
Alexandre Klementiev, Ivan Titov, and Binod Bhattarai. 2012.
\newblock Inducing crosslingual distributed representations of words.
\newblock \emph{Proceedings of the 24th International Conference on
  Computational Linguistics}.

\bibitem[{Kobdani et~al.(2009)Kobdani, Fraser, and Sch\"{u}tze}]{kobdani09word}
Hamidreza Kobdani, Alex Fraser, and Hinrich Sch\"{u}tze. 2009.
\newblock Word alignment by thresholded two-dimensional normalization.
\newblock In \emph{Proceeedings of the 12th Machine Translation Summit}.

\bibitem[{Lardilleux and Lepage(2007)}]{lardilleux2007contribution}
Adrien Lardilleux and Yves Lepage. 2007.
\newblock The contribution of the notion of hapax legomena to word alignment.
\newblock In \emph{Proceedings of the 4th Language and Technology Conference}.

\bibitem[{Lardilleux and Lepage(2009)}]{lardilleux2009sampling}
Adrien Lardilleux and Yves Lepage. 2009.
\newblock Sampling-based multilingual alignment.
\newblock In \emph{Proceedings of 7th Conference on Recent Advances in Natural
  Language Processing}.

\bibitem[{Levy et~al.(2017)Levy, S{\o}gaard, and Goldberg}]{levy2017strong}
Omer Levy, Anders S{\o}gaard, and Yoav Goldberg. 2017.
\newblock A strong baseline for learning cross-lingual word embeddings from
  sentence alignments.
\newblock In \emph{Proceedings of the 15th Conference of the European Chapter
  of the Association for Computational Linguistics}.

\bibitem[{Luong et~al.(2015)Luong, Pham, and Manning}]{luong2015bilingual}
Thang Luong, Hieu Pham, and Christopher~D Manning. 2015.
\newblock Bilingual word representations with monolingual quality in mind.
\newblock In \emph{Proceedings of the 1st Workshop on Vector Space Modeling for
  Natural Language Processing}.

\bibitem[{Malaviya et~al.(2017)Malaviya, Neubig, and
  Littell}]{malaviya17typology}
Chaitanya Malaviya, Graham Neubig, and Patrick Littell. 2017.
\newblock Learning language representations for typology prediction.
\newblock In \emph{Proceedings of the 2017 Conference on Empirical Methods in
  Natural Language Processing}.

\bibitem[{Mayer and Cysouw(2014)}]{mayer2014creating}
Thomas Mayer and Michael Cysouw. 2014.
\newblock Creating a massively parallel bible corpus.
\newblock In \emph{Proceedings of the 9th International Conference on Language
  Resources and Evaluation}.

\bibitem[{McDonald et~al.(2011)McDonald, Petrov, and
  Hall}]{mcdonald11delexicalized}
Ryan~T. McDonald, Slav Petrov, and Keith~B. Hall. 2011.
\newblock Multi-source transfer of delexicalized dependency parsers.
\newblock In \emph{Proceedings of the 2011 Conference on Empirical Methods in
  Natural Language Processing}.

\bibitem[{Melamed(1997)}]{melamed97wordtoword}
I.~Dan Melamed. 1997.
\newblock A word-to-word model of translational equivalence.
\newblock In \emph{Proceedings of the 35th Annual Meeting of the Association
  for Computational Linguistics and 8th Conference of the European Chapter of
  the Association for Computational Linguistics}.

\bibitem[{Mikolov et~al.(2013{\natexlab{a}})Mikolov, Chen, Corrado, and
  Dean}]{mikolov2013efficient}
Tomas Mikolov, Kai Chen, Greg Corrado, and Jeffrey Dean. 2013{\natexlab{a}}.
\newblock Efficient estimation of word representations in vector space.
\newblock \emph{arXiv preprint arXiv:1301.3781}.

\bibitem[{Mikolov et~al.(2013{\natexlab{b}})Mikolov, Le, and
  Sutskever}]{mikolov2013exploiting}
Tomas Mikolov, Quoc~V Le, and Ilya Sutskever. 2013{\natexlab{b}}.
\newblock Exploiting similarities among languages for machine translation.
\newblock \emph{arXiv preprint arXiv:1309.4168}.

\bibitem[{Och and Ney(2003)}]{och03alignment}
Franz~Josef Och and Hermann Ney. 2003.
\newblock A systematic comparison of various statistical alignment models.
\newblock \emph{Computational Linguistics}, 29(1).

\bibitem[{{\"O}stling(2014)}]{ostling2014bayesian}
Robert {\"O}stling. 2014.
\newblock Bayesian word alignment for massively parallel texts.
\newblock In \emph{Proceedings of the 14th Conference of the European Chapter
  of the Association for Computational Linguistics}.

\bibitem[{{\"O}stling and Tiedemann(2016)}]{ostling2016efmaral}
Robert {\"O}stling and J{\"o}rg Tiedemann. 2016.
\newblock Efficient word alignment with {M}arkov {C}hain {M}onte {C}arlo.
\newblock \emph{Prague Bulletin of Mathematical Linguistics}, 106.

\bibitem[{{\"O}stling and Tiedemann(2017)}]{ostling2017continuous}
Robert {\"O}stling and J{\"o}rg Tiedemann. 2017.
\newblock Continuous multilinguality with language vectors.
\newblock In \emph{Proceedings of the 15th Conference of the European Chapter
  of the Association for Computational Linguistics}.

\bibitem[{{Sarath Chandar} et~al.(2014){Sarath Chandar}, Lauly, Larochelle,
  Khapra, Ravindran, Raykar, and Saha}]{ap2014autoencoder}
AP~{Sarath Chandar}, Stanislas Lauly, Hugo Larochelle, Mitesh~M. Khapra,
  Balaraman Ravindran, Vikas~C. Raykar, and Amrita Saha. 2014.
\newblock An autoencoder approach to learning bilingual word representations.
\newblock In \emph{Proceedings of the 2014 Annual Conference on Neural
  Information Processing Systems}.

\bibitem[{Simard(1999)}]{simard1999text}
Michel Simard. 1999.
\newblock Text-translation alignment: Three languages are better than two.
\newblock In \emph{Proceedings of the 1999 Joint SIGDAT Conference on Empirical
  Methods in Natural Language Processing and Very Large Corpora}.

\bibitem[{S{\o}gaard et~al.(2015)S{\o}gaard, Agi{\'c}, Alonso, Plank, Bohnet,
  and Johannsen}]{sogaard2015inverted}
Anders S{\o}gaard, {\v{Z}}eljko Agi{\'c}, H{\'e}ctor~Mart{\'\i}nez Alonso,
  Barbara Plank, Bernd Bohnet, and Anders Johannsen. 2015.
\newblock Inverted indexing for cross-lingual nlp.
\newblock In \emph{The 53rd Annual Meeting of the Association for Computational
  Linguistics and the 7th International Joint Conference of the Asian
  Federation of Natural Language Processing}.

\bibitem[{Somers(2005)}]{somers2005round}
Harold Somers. 2005.
\newblock Round-trip translation: What is it good for?
\newblock In \emph{Proceedings of the Australasian Language Technology Workshop
  2005}.

\bibitem[{Swadesh(1946)}]{swadesh1971south}
Morris Swadesh. 1946.
\newblock South {G}reenlandic ({E}skimo).
\newblock In Cornelius Osgood, editor, \emph{Linguistic Structures of Native
  America}. Viking Fund Inc. (Johnson Reprint Corp.), New York.

\bibitem[{Tiedemann(2003)}]{tiedemann03clues}
J{\"{o}}rg Tiedemann. 2003.
\newblock Combining clues for word alignment.
\newblock In \emph{Proceedings of the 10th Conference of the European Chapter
  of the Association for Computational Linguistics}.

\bibitem[{Tiedemann(2018)}]{tiedemann2018emerging}
J{\"o}rg Tiedemann. 2018.
\newblock Emerging language spaces learned from massively multilingual corpora.
\newblock \emph{arXiv preprint arXiv:1802.00273}.

\bibitem[{Tsvetkov et~al.(2014)Tsvetkov, Boytsov, Gershman, Nyberg, and
  Dyer}]{tsvetkov14metaphor}
Yulia Tsvetkov, Leonid Boytsov, Anatole Gershman, Eric Nyberg, and Chris Dyer.
  2014.
\newblock Metaphor detection with cross-lingual model transfer.
\newblock In \emph{Proceedings of the 52nd Annual Meeting of the Association
  for Computational Linguistics}.

\bibitem[{Upadhyay et~al.(2016)Upadhyay, Faruqui, Dyer, and
  Roth}]{upadhyay2016cross}
Shyam Upadhyay, Manaal Faruqui, Chris Dyer, and Dan Roth. 2016.
\newblock Cross-lingual models of word embeddings: An empirical comparison.
\newblock In \emph{Proceedings of the 54th Annual Meeting of the Association
  for Computational Linguistics}.

\bibitem[{Vuli{\'c} and Moens(2012)}]{vulic2012detecting}
Ivan Vuli{\'c} and Marie-Francine Moens. 2012.
\newblock Detecting highly confident word translations from comparable corpora
  without any prior knowledge.
\newblock In \emph{Proceedings of the 13th Conference of the European Chapter
  of the Association for Computational Linguistics}.

\bibitem[{Vulic and Moens(2012)}]{vulic2012sub}
Ivan Vulic and Marie-Francine Moens. 2012.
\newblock Sub-corpora sampling with an application to bilingual lexicon
  extraction.
\newblock In \emph{Proceedings of the 24th International Conference on
  Computational Linguistics}.

\bibitem[{Vuli{\'c} and Moens(2015)}]{vulic2015bilingual}
Ivan Vuli{\'c} and Marie-Francine Moens. 2015.
\newblock Bilingual word embeddings from non-parallel document-aligned data
  applied to bilingual lexicon induction.
\newblock In \emph{Proceedings of the 53rd Annual Meeting of the Association
  for Computational Linguistics and the 7th International Joint Conference on
  Natural Language Processing}, volume~2.

\bibitem[{Wang et~al.(2016)Wang, Zhao, Ploux, Lu, Utiyama, and
  Sumita}]{wang2016novel}
Rui Wang, Hai Zhao, Sabine Ploux, Bao-Liang Lu, Masao Utiyama, and Eiichiro
  Sumita. 2016.
\newblock A novel bilingual word embedding method for lexical translation using
  bilingual sense clique.
\newblock \emph{arXiv preprint arXiv:1607.08692}.

\bibitem[{Xiao and Guo(2014)}]{xiao2014distributed}
Min Xiao and Yuhong Guo. 2014.
\newblock Distributed word representation learning for cross-lingual dependency
  parsing.
\newblock In \emph{Proceedings of the 18th Conference on Computational Natural
  Language Learning}.

\bibitem[{Zeman and Resnik(2008)}]{zeman08crosslanguage}
Daniel Zeman and Philip Resnik. 2008.
\newblock Cross-language parser adaptation between related languages.
\newblock In \emph{Proceedings of the 3rd International Joint Conference on
  Natural Language Processing}.

\bibitem[{Zou et~al.(2013)Zou, Socher, Cer, and Manning}]{zou2013bilingual}
Will~Y Zou, Richard Socher, Daniel Cer, and Christopher~D Manning. 2013.
\newblock Bilingual word embeddings for phrase-based machine translation.
\newblock In \emph{Proceedings of the 2013 Conference on Empirical Methods in
  Natural Language Processing}.

\end{thebibliography}
\bibliographystyle{acl_natbib}

\end{document}